\documentclass[
]{ceurart}

\usepackage{listings}
\usepackage{hyperref}
\usepackage{tabularx}
\usepackage{subcaption}
\begin{document}

\copyrightyear{2023}
\copyrightclause{Copyright for this paper by its authors.
  Use permitted under Creative Commons License Attribution 4.0
  International (CC BY 4.0).}

\conference{CLEF 2023: Conference and Labs of the Evaluation Forum, September 18–21, 2023, Thessaloniki, Greece}

\title{SuryaKiran at MEDIQA-Sum 2023: Leveraging LoRA for Clinical Dialogue Summarization}



\author{Kunal Suri}[%
email=kunal_suri@optum.com,
]
\cormark[1]

\author{Prakhar Mishra}[%
email=prakhar_mishra29@optum.com,
]
\fnmark[1]

\author{Saumajit Saha}[%
email=saumajit_saha@optum.com,
]
\fnmark[1]

\author{Atul Singh}[%
email=atul_singh18@optum.com,
]
\address{Optum, India}

\cortext[1]{Corresponding author.}
\fntext[1]{These authors contributed equally.}

\begin{abstract}
Finetuning Large Language Models helps improve the results for domain-specific use cases. End-to-end finetuning of large language models is time and resource intensive and has high storage requirements to store the finetuned version of the large language model. Parameter Efficient Fine Tuning (PEFT) methods address the time and resource challenges by keeping the large language model as a fixed base and add additional layers, which the PEFT methods finetune. This paper demonstrates the evaluation results for one such PEFT method Low Rank Adaptation (LoRA), for Clinical Dialogue Summarization. The evaluation results show that LoRA works at par with end-to-end finetuning for a large language model. The paper presents the evaluations done for solving both the Subtask A and B from ImageCLEFmedical \footnote{https://www.imageclef.org/2023/medical}
\end{abstract}


\maketitle

\section{Introduction}

It is important to record conversations between medical personnel and patients for compliance, training, and evaluation purposes. To that end, summaries of such conversations serve as valuable tools for medical personnel and patients to refer back to and comprehend their prior interactions. Therefore, a concise summary must be produced to facilitate the next medical consultation and provide a source for future reference. Currently, such summaries are created manually; this summarization process is costly and labour-intensive. AI-based summarization techniques can help here by reducing the time and cost associated with manual summarization and facilitating the generation of more accurate representations of doctor-patient conversations by human scribes in less time.

Sequence-to-Sequence (Seq2Seq) Architectures \citep{NIPS2014_a14ac55a} have been at the forefront of creating summaries. Transformers \citep{NIPS2017_3f5ee243} further improved the performance of this architecture. Over time, we have seen that the performance of these models have improved significantly \footnote{https://paperswithcode.com/sota/text-summarization-on-pubmed-1} but it comes at the cost of increased model size which made it very difficult to fit such models on consumer grade hardware such as K80 or T4. Recently a couple of techniques such as LoRA \citep{hu2021lora}, Prefix Tuning \citep{li2021prefixtuning}, P-Tuning \citep{liu2021gpt}, Prompt Tuning \citep{lester-etal-2021-power} have been introduced which are collectively referred to as \textit{Parameter Efficient Fine Tuning} (PEFT) techniques. These techniques are used for efficiently adapting pre-trained language models (PLMs) to various downstream applications without fine-tuning all the model’s parameters. PEFT methods only trains a small number of (extra) model parameters, significantly decreasing computational and storage costs because fine-tuning large-scale PLMs is prohibitively costly. For this paper, we use the PEFT implementation from Huggingface \footnote{https://huggingface.co/docs/peft/index}.

This paper presents the experimental results of our explorations with LoRA on Clinical Dialogs to accomplish both Subtask A and B of \citep{MEDIQA-Sum2023} Shared Tasks from \citep{ImageCLEF2023}. The solution of SubTask B presented in this paper was ranked first among all the submissions for SubTask B. The paper uses LoRA based models for both assigning conversations to a pre-defined set of clinical notes sections and summarization of conversations. Through this work, the paper also compares the performance of fine-tuned Transformer based models with LoRA based models for classification and summarization tasks. In addition to this comparison, we also evaluate impact of ensembling outputs from multiple Seq2Seq models using \citep{kobayashi-2018-frustratingly}. Our simulations show that LoRA works as well as finetuning of Transformer-based models. This is very important because it shows that we can get the equivalent performance as we get after fine tuning Transformer models while using only a fraction of parameters which means that such models could be fine tuned on consumer grade hardware such as K80 and T4.

This paper is organized as follows. Section \ref{sec:imageclef_task} presents a brief overview of SubTask A and B - including available labeled data and evaluation metrics. Then the paper describes current state-of-the-art for dialog classification and summarization in Section \ref{sec:related_work} that this paper builds upon. This is followed by the description of the approach used to solve SubTask A in Section \ref{sec:task_a_approach} and SubTask B in Section \ref{sec:task_b_approach}. Then the results of our solutions for both of these subtasks are presented. Finally, the paper ends with a conclusion of the work. The paper includes an appendix containing exploratory data analysis and material that will help to better understand the solution presented in the paper.

\section{Related Work} \label{sec:related_work}

Finetuning large language models enables better performance for domain-specific use cases. In-context finetuning performs well in few-shot scenarios enabling the end users to provide examples with the prompt to enable LLMs to learn for the use case at hand. This approach does not scales as it restricts sending multiple examples with the prompt. End-to-end finetuning of LLMs is resource and time intensive and has the additional drawback of storing and managing multiple copies of large-size models.

Parameter Efficient Fine Tuning (PEFT) Methods attempt to solve the problems mentioned above by finetuning a smaller number of existing or newly introduced parameters of the large language model while keeping the rest of the parameters frozen. In  \cite{lialin2023scaling}, Lilian et al. divide PEFT methods into the following four categories: additive, selective, reparameterization-based, and hybrid methods. Additive methods such as adapters \cite{DBLP:journals/corr/abs-1902-00751} introduce and train only a new set of parameters or layers. Selective methods finetune only a few top layers of the network. Reparametrization-based methods use a low-dimensional representation of the network to reduce the number of parameters to be trained during finetuning. This paper evaluates Low-Rank Adaptation (LoRA) a prominent example of this category of methods.

Parameter Efficient Fine Tuning (PEFT) methods reduce the need to host a large-sized model for each use case. They enable users to use a frozen base model with a small layer of model weights that vary with the use case. In \cite{pu2023empirical}, the authors compare the performance of four different PEFT techniques for scenarios where low, medium and high counts of samples are available for fine-tuning. The evaluation results show that LoRA gives near-best performance when low to medium data samples are available for summarization tasks. In another similar related study in \cite{vanveen2023radadapt}, the evaluations demonstrate that the best summarization for radiology reports is achieved using a model  pre-trained on the clinical text and then fine-tuned using LoRA. In this paper, the authors have used LoRA and ensembling for summarization.

\section{Task Description} \label{sec:imageclef_task}
This Section provides a high-level overview of the MEDIQA-Sum 2023 Task (including both SubTask A and B) from ImageCLEFmed MEDIQA\citep{ImageCLEF2023}. The Section starts with a description of different SubTask goals followed by basic counts of available labeled data. The metric used to evaluate this task is arithmetic mean of ROUGE-1 \citep{lin-2004-rouge}, Bertscore F1 \citep{DBLP:journals/corr/abs-1904-09675}, and BLEURT \citep{sellam2020bleurt}.

\subsection{Task Definition} \label{sec:imageclef_task_definition}

Given a short conversation between a Doctor and a patient or another Doctor (\textbf{Dialogue}), the goal of SubTask A is to create a system that automatically predicts the Section to which the conversation belongs to which is denoted by \textbf{Section Header}. There are twenty Sections Headers in this dataset. Some examples of Section Headers are FAM/SOCHX, GENHX, PASTMEDICALHX, CC. All of these Section Headers and their descriptions (\textbf{Section Description}) can be found in Table \ref{tab:taska_sec_header}. The goal of SubTask B is to create a system that generates a summary which matches the human generated summary (\textbf{Section Text}) as closely as possible while optimizing the metric for evaluation.

\subsection{Labeled Data} \label{sec:imageclef_task_labelled_data}

In this paper we have used the labeled data provided by MEDIQA-Sum 2023 organizers for training the models. A sample data point from the labeled data set for SubTask A and B can be found in Table \ref{tab:taskA_sample}. The official data consists of a training and validation split. For SubTask A and B, training data contains 1201 and validation data contains 180 <dialogue, section-text, section-header> triplets.

\section{SubTask A Methodology} 
\label{sec:task_a_approach}
\begin{figure*}[h]
\centering
\caption{SubTask A - Overall Architecture}
\label{fig:taska_overall_architecture}
\includegraphics[width=0.8\textwidth, height = 7.5cm]{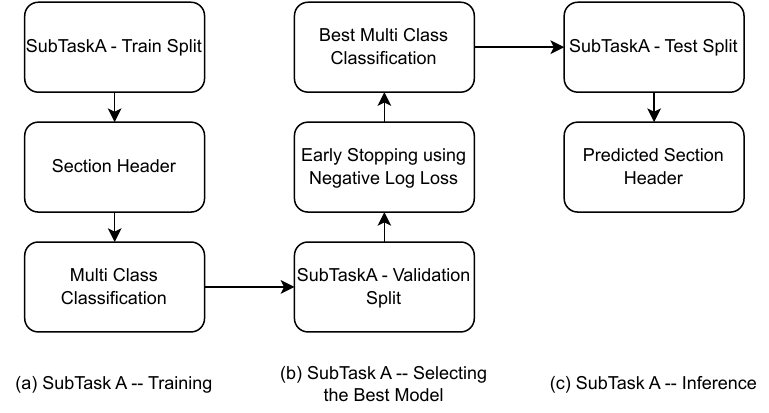}
\end{figure*}

Given a short conversation between a doctor and a patient, the goal of SubTask A is to predict its Section Header. This Section starts with a description of the approach used to predict the Section Header. 

We have achieved success using Bio-ClinicalBERT \citep{DBLP:journals/corr/abs-1904-03323} for classification in the healthcare domain. Hence we choose it as the backbone and initialize LoRA layer on top of it. We use this architecture for classification of Dialogue to a Section Header in SubTask A. We limit the number of input tokens to 300 tokens because that is the length of majority of dialogues, as shown in Figure \ref{fig:taska_section_header_class_distribution}. We use a 3 Fold Cross Validation approach for modeling purposes. This is to ensure that we capture all information in the data. For every fold, we split its test part into validation and test. We do this so that we can use validation split to select best model using Early Stopping and test split to calculate its performance. The hyper-parameters used for training and performance for all folds can be found in Table \ref{tab:taska_section_header_prediction}. During inference, we pass a given Dialogue through all three models, take an average of the logits for all the classes and output the class with the highest logit score.

\section{SubTask B Methodology} \label{sec:task_b_approach}

\begin{figure*}[h]
\centering
\caption{SubTask B - Overall Architecture}
\label{fig:taskb_overall_architecture}
\includegraphics[width=0.8\textwidth, height = 7.5cm]{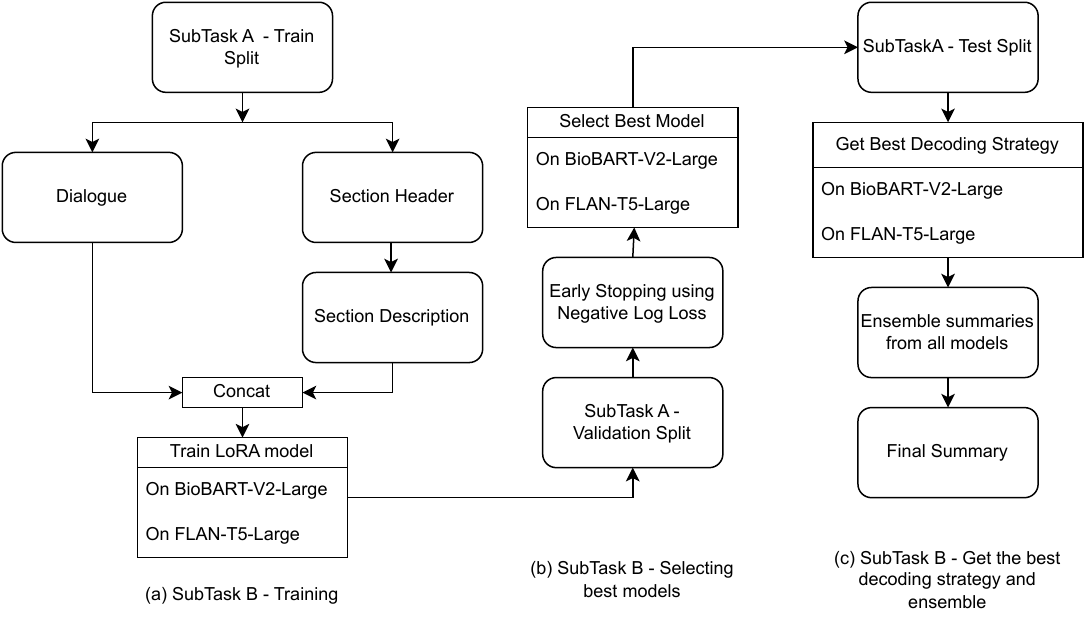}
\end{figure*}

Given a short conversation between a doctor and a patient, the goal of SubTask B is to summarize it while ensuring that the generated summary is as fluent and as close to Section Text as possible. This Section starts with a description of the methodology used to summarize the conversation. For Dialogue Summarization, we have trained a LoRA layer on top of Seq2Seq models. This Section also describes the processed labeled data used for training these models, followed by the actual training steps. Then this Section looks at the steps used to generate the summary from the decoder. Finally, we discuss the approach used for ensembling the outputs of these models.

We train LoRA based Seq2Seq models using labeled data (Dialogue + Section Header, Section Text) as (Input, Output) pair. Section Text is a part of the labeled data and is a human subject matter expert-created summary of Dialogue. As a preprocessing step, we replace all new line characters with whitespaces. The Dialogue is concatenated with the section description of its Section Header by the SEP token of the Seq2Seq architecture. During training and inference, we use the actual section description for the actual Section Header. No changes are made to Section Text.

We use a 3-fold cross validation scheme as described in \ref{sec:task_a_approach} and train LoRA on two Seq2Seq architectures - BioBart-V2-Large \cite{yuan2022biobart} and Flan-T5-Large \cite{chung2022scaling}. Here we need to select the number of input tokens for encoder and decoder. For encoder, we have selected token length of 512 tokens and for decoder, we have selected token length of 400 tokens. All the hyper-parameters used to train each of the above architecture can be found in Table \ref{tab:taska_seq2seq_finetuning_hp}. To select the best model, we use early-stopping \cite{yao2007early} based on Validation Negative Log Loss. Results on the test part of each of these models can be found in Table \ref{tab:taskb_summary_comparison}. The distribution of tokens for Dialogue and Section Text can be found in Figure \ref{fig:taska_overall_dialogue_token_distribution} and Figure \ref{fig:taska_overall_notes_token_distribution} respectively.

To generate summaries that match the human generated summaries, we need a way to control the text generated by the decoder component of a Seq2Seq model. This can be done by using decoding strategies such as Beam Search \citep{DBLP:journals/corr/abs-1211-3711}, Top-k Sampling \citep{fan-etal-2018-hierarchical}, Top-p Sampling \citep{DBLP:journals/corr/abs-1904-09751}, Contrastive Search \citep{su2023contrastive} etc. In this module, we use Beam Search with TPESampler Algorithm from Optuna\footnote{\url{https://optuna.readthedocs.io/en/stable/reference/samplers/generated/optuna.samplers.TPESampler.html}} to search for the optimal decoding strategy trying to maximize ROUGE-1, ROUGE-2, and BertScore rather than relying on manual tweaking of these metrics. We use TPESampler here because it supports multivariate optimization and also it handles Float, Integer, and Categorical values better than other algorithms present in Optuna\footnote{\url{https://optuna.readthedocs.io/en/stable/reference/samplers/index.html}}. We use Optuna here due to ease of implementing Hyper-parameter optimization algorithms. We did not use BLEURT during search because it is extremely time consuming. For this module, we use four hyper-parameters for Beam Search - Early Stopping, Number of Beams, No Repeat N-gram Size, Length Penalty. The search space of each of these variables can be found in the Table \ref{tab:task_a_search_space}.

The results from the different models are ensembled using \textbf{Generating Best Summary by semantic similarity} \label{taska_best_summary_ensemble}  - a post-ensemble method \cite{kobayashi-2018-frustratingly} to identify the summary which is closest to all the generated summaries. The paper uses this output summary as the final summary for the given Dialogue.

\begin{table}[!ht]
\centering
\caption{Search Space for Beam Search Decoding}
\label{tab:task_a_search_space}
\begin{tabular}{|c|c|c|}
\hline
\textbf{Variable} & \textbf{Data Type} & \textbf{Range} \\
\hline
Early Stopping & Categorical & [True,False] \\
\hline
Number of Beams & Integer & 5-15 \\
\hline
No Repeat Ngram Size & Integer & 5-15 \\
\hline
Length Penalty & Float & [-2,2] \\
\hline
\end{tabular}
\end{table}

\section{SubTask A Results and Analysis} \label{task_a_experiments_and_results}

This Section presents the results for SubTask A using the approach described in Section \ref{sec:task_a_approach}.
We have made only one submission for predicting Section Header whose Multi Class Accuracy was 73.5\% on the test set given by the organizers, obtaining a rank of 8 among 23 submissions. In this submission, we pass Dialogues through all three LoRA based Bio-ClinicalBERT models, take an average of the logits for all the classes and output the class with the highest logit score. The table containing our team's standing can be found in the Tables \ref{tab:taska_section_header_classification_standings}. Standings of all the teams have been calculated using multi class accuracy. We compared performance of Bio-ClinicalBERT when it is fine-tuned end-to-end and when it is used as a backbone for LoRA. We observe that Bio-ClinicalBERT with LoRA score 73.3\% on validation data whereas end-to-end fine-tuned Bio-ClinicalBERT score 72\% on the  same validation data.

\section{SubTask B Results} \label{task_b_experiments_and_results}

This Section presents the results for SubTask B using the approach described in Section \ref{sec:task_b_approach}.
We have made three submissions (mentioned as \textit{runs} in the result tables) for generating summaries from Dialogues. For the summarization task, we have submitted results from three \textit{runs}. In \textit{run 1} and \textit{run 2}, we train LoRA on BioBart-V2-Large and Flan-T5-Large respectively while \textit{run 3} presents the results of ensembling summaries from both of these models. The details for each run are as follows:

\begin{enumerate}
  \item Run 1 - We generate summary from BioBart-V2-Large model trained on each fold and ensemble output of all the models using  \ref{taska_best_summary_ensemble} 
  \item Run 2 - We generate summary from Flan-T5-Large model trained on each fold and ensemble output of all the models using \textit{Generating Best Summary by semantic similarity}.
  \item Run 3 - We generate summary from BioBart-V2-Large and Flan-T5-Large model trained on each fold and ensemble output of all the models using \textit{Generating Best Summary by semantic similarity}.
\end{enumerate}

The table containing our team's standing can be found in Table \ref{tab:taskb_section_text_summarization_standings}. Standings of all the teams have been calculated by calculating arithmetic mean of Rouge-1, Bertscore, BLEURT for the Dialogue summary.

The experiments show that Run3 performs the best scoring rank 1 out of 13 submissions. This is also intuitive since it contains summaries from 3 models of BioBART-V2-Large and 3 models of Flan-T5-Large. Run2 scored 5th rank and Run1 scored 6th rank. This is an interesting observation since Flan-T5-Large is an enhanced version of T5 that has been finetuned in a mixture of tasks whereas BioBart-V2-Large has been trained solely on medical corpus so ideally Run1 should have scored better than Run2 but it seems that bigger models work better than domain specific models although this hypothesis needs to be validated. 

\begin{table}[!ht]
    \centering
    \caption{Results of runs on Test Data}
    \label{tab:taskb_summary_run_results_comparisons}
    \begin{tabular}{|c|c|c|c|c|}
    \hline
    \textbf{Run} & \textbf{ROUGE-1} & \textbf{Bertscore-F1}  & \textbf{BLEURT} & \textbf{Mean Score} \\
    \hline
    Run 3 & 0.4398 & 0.7231 & 0.5567 & 0.5732 \\
    \hline
    Run 2 & 0.4209 & 0.7137 & 0.5423 & 0.5590 \\
    \hline
    Run 1 & 0.4056 & 0.7109 & 0.5324 & 0.5496 \\
    \hline
    \end{tabular}
\end{table}

\subsection{Analysis of different Transformer Architectures on SubTask B}

We compare performance of BioBart-V2-Large and Flan-T5-Large when they are fine-tuned end-to-end and they are treated as backbone for LoRA. We observe that the models trained with LoRA perform better than the models which were fine-tuned end-to-end. The performance was evaluated by calculating arithmetic mean of ROUGE-1, ROUGE-2, and BertScore-F1. We do not use BLEURT here as it is extremely time consuming and based on our observations, ROUGE-2 and BLEURT have a very strong correlation. The average score across all folds for each architecture can be found in the Table \ref{tab:taskb_summary_comparison}.

\section{Conclusion}

The paper presents the solution and the results for SubTask A and B of ImageCLEFmed MEDIQA-Sum task. The solution uses LoRA to finetune Transformer based models to classify and summarise Clinical Dialogues, and our simulation results show that the performance of Transformer based models finetuned using LoRA is equivalent to the performance of Transformer based models finetuned using resource and time-intensive end-to-end finetuning. The success of Transformer based model finetunes using LoRA implies organizations can easily finetune and deploy domain-based models. 

The authors observe that metrics such as ROUGE are ineffective for evaluating the performance of models like OpenAI GPT3 as they focus on syntactic similarity. Metrics such as Bertscore and BLEURT seem more suitable for such models since they focus on semantic similarity. Finally, the paper also evaluates two different ensemble techniques, and the results demonstrate that the Post Ensemble technique performs the best while giving minimum hallucinations.

\appendix
\setcounter{table}{0}
\renewcommand{\thetable}{A\arabic{table}}
\setcounter{figure}{0}
\renewcommand{\thefigure}{A\arabic{figure}}
\clearpage
\section{Appendix}
\subsection{Data Exploration and Explanation}
This section discusses data exploration and explanation so that audience can understand why we made the decisions that we made.
A sample data point from dataset for SubTask A and B can be seen in Table \ref{tab:taskA_sample}.
\begin{table}[h]
\centering
\caption{Sample data point for SubTask A and B}
\label{tab:taskA_sample}
\begin{tabularx}{\linewidth}{|c|X|}
\hline
\textbf{Variable} & \textbf{Sample Value} \\ \hline
Section Header & FAM/SOCHX \\ \hline
Section Text & The patient has been a smoker since the age of 10. So, he was smoking 2-3 packs per day. Since being started on Chantix, he says he has cut it down to half a pack per day. He does not abuse alcohol \\ \hline
Dialogue & Doctor: Are you a smoker? \newline Patient: Yes. I do not drink if that is any constellation. \newline Doctor: How much do you smoke per day? \newline Patient: I just started taking Chantix and now I am down to a half a pack a day. \newline Doctor: How much did you smoke per day prior to starting Chantix? \newline Patient: I was smoking about two to three packs a day. I have been smoker since I was ten years old. \\ \hline
\end{tabularx}
\end{table}

The description of each of the Section Headers present in the data can be found in Table \ref{tab:taska_sec_header}
\begin{table}[h]
\centering
\caption{Section Headers and their descriptions.}
\label{tab:taska_sec_header}
\begin{tabularx}{\linewidth}{|c|X|} 
\hline
\textbf{Section Header} & \textbf{Section Header Description} \\ \hline
FAM/SOCHX & FAMILY HISTORY/SOCIAL HISTORY \\ \hline
GENHX & HISTORY OF PRESENT ILLNESS \\ \hline
PASTMEDICALHX & PAST MEDICAL HISTORY \\ \hline
CC & CHIEF COMPLAINT \\ \hline
PASTSURGICAL & PAST SURGICAL HISTORY \\ \hline
ALLERGY & ALLERGY \\ \hline
ROS & REVIEW OF SYSTEMS \\ \hline
MEDICATIONS & MEDICATIONS \\ \hline
ASSESSMENT & ASSESSMENT \\ \hline
EXAM & EXAM \\ \hline
DIAGNOSIS & DIAGNOSIS \\ \hline
DISPOSITION & DISPOSITION \\ \hline
PLAN & PLAN \\ \hline
EDCOURSE & EMERGENCY DEPARTMENT COURSE \\ \hline
IMMUNIZATIONS & IMMUNIZATIONS \\ \hline
IMAGING & IMAGING \\ \hline
GYNHX & GYNECOLOGIC HISTORY \\ \hline
PROCEDURES & PROCEDURES \\ \hline
OTHER$\_$HISTORY & OTHER$\_$HISTORY \\ \hline
LABS & LABS \\ \hline
\end{tabularx}
\end{table}

\begin{figure}
    \centering
    \caption{Class distribution of Section Headers}
    \label{fig:taska_section_header_class_distribution}
    \includegraphics[width=\textwidth]{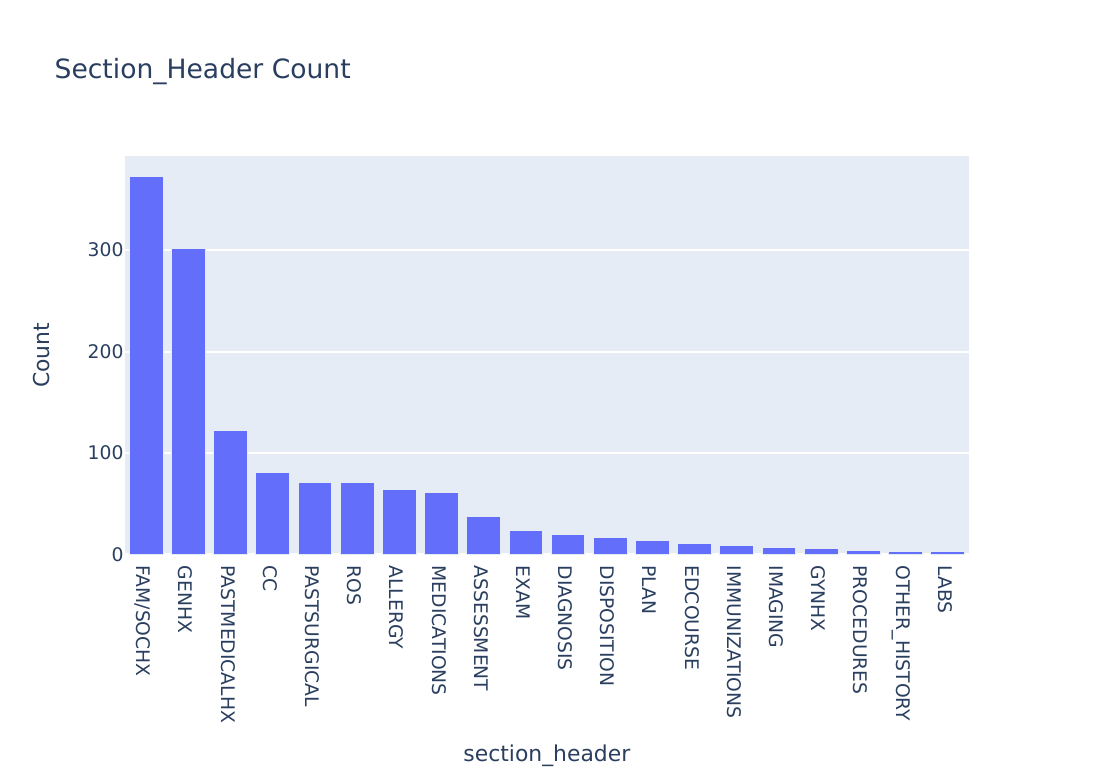}
\end{figure}

The Class distribution of Section Headers for SubTask A is give by Figure \ref{fig:taska_section_header_class_distribution}

\begin{figure}
    \centering
    \caption{Dialogue Token Distribution}
    \label{fig:taska_overall_dialogue_token_distribution}
    \includegraphics[width=\textwidth]{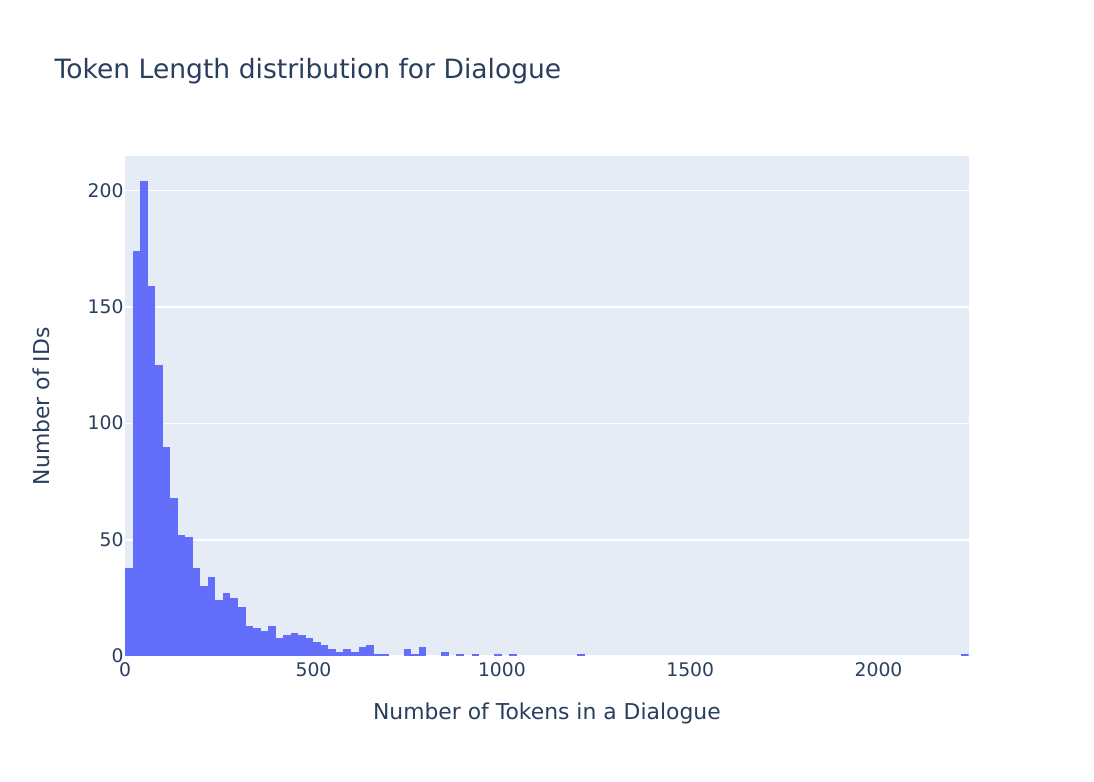}
\end{figure}

The Dialogue Token Distribution for SubTask A and B is give by Figure \ref{fig:taska_overall_dialogue_token_distribution}

\begin{figure}
    \centering
    \caption{Clinical Note Token Distribution}
    \label{fig:taska_overall_notes_token_distribution}
    \includegraphics[width=\textwidth]{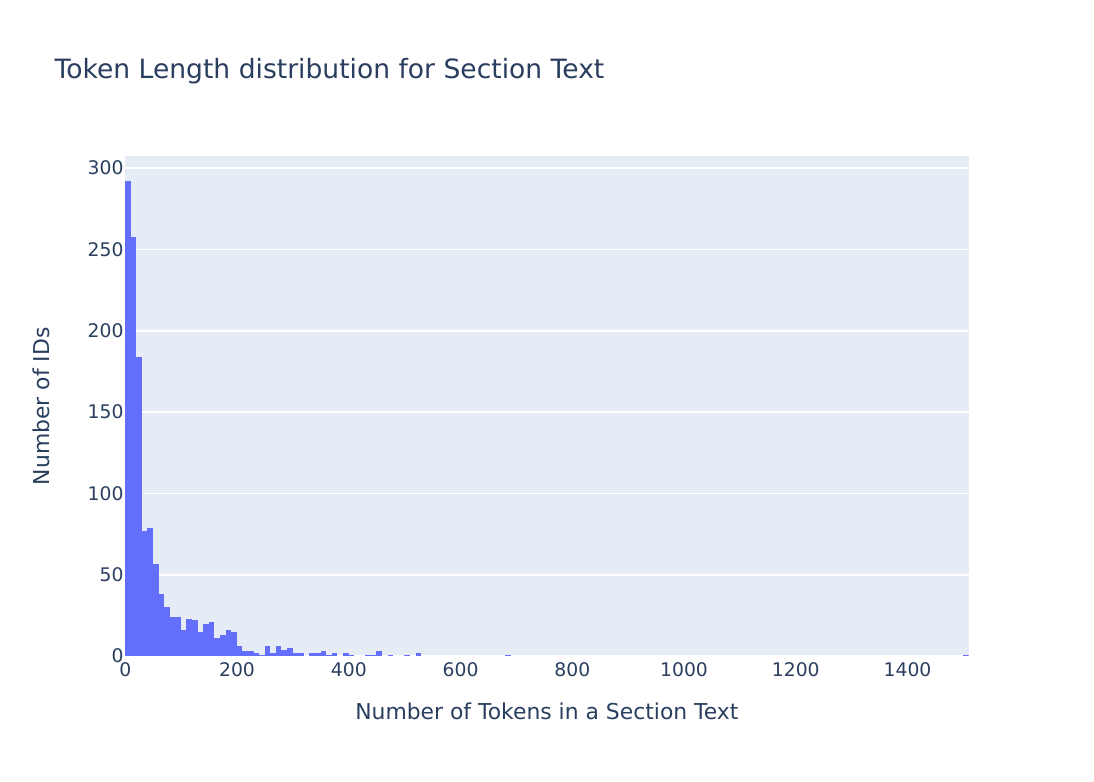}
\end{figure}

The Clinical Note Token Distribution for SubTask B is give by Figure \ref{fig:taska_overall_notes_token_distribution}

The hyper-parameters and performance metrics for Predicting Section Header i.e SubTask A can be found in the Table \ref{tab:taska_section_header_prediction}. 

\begin{table*}[h]
\centering
\caption{SubTask A - Predicting Section Header. Base Arch: Base Architecture, BS: Batch Size, LR: Learning Rate, LoRA-A: LoRA-Alpha, LoRA-D: LoRA-Dropout  BVL : Best Validation Loss.}
\label{tab:taska_section_header_prediction}
\begin{tabular}{|c|c|c|c|c|c|c|c|c|}
\hline
\textbf{Base Arch} & \textbf{Fold} & \textbf{Epochs} & \textbf{BS} & \textbf{LR} & \textbf{LoRA-R} & \textbf{LoRA-A} & \textbf{LoRA-D} & \textbf{BVL} \\
\hline
Bio-ClinicalBERT & 0 & 150 & 16 & 1e-3 & 8 & 32 & 0.01 & 1.193 \\
\hline
Bio-ClinicalBERT & 1 & 150 & 16 & 1e-3 & 8 & 32 & 0.01 & 1.429 \\
\hline
Bio-ClinicalBERT & 2 & 150 & 16 & 1e-3 & 8 & 32 & 0.01 & 0.4961 \\
\hline
\end{tabular}
\end{table*}

The hyperparameters used to fine tune Seq2Seq Models and LoRA i.e. SubTask B can be found in Table \ref{tab:taska_seq2seq_finetuning_hp}. Each of these models were trained on 150 epochs, Gradient Accumulation of 16, Learning rate of 1e-3, AdamW optimizer, and Linear Learning Scheduler.
\begin{table*}[h]
\centering
\caption{SubTask B - Hyperparameter Tuning for Different Architectures. Base Arch: Base Architecture, BS: Batch Size, LR : Learning Rate, LoRA-A: LoRA-Alpha, LoRA-D: LoRA-Dropout, MaxSL : Maximum Source Length, MaxTL : Maximum Target Length, MinTL : Minimum Target Length}
\label{tab:taska_seq2seq_finetuning_hp}
\begin{tabular}{|c|c|c|c|c|c|c|c|c|}
\hline
\textbf{Base Arch} & \textbf{BS} & \textbf{LR} & \textbf{LoRA-R} & \textbf{LoRA-A} & \textbf{LoRA-D} & \textbf{MaxSL} & \textbf{MaxTL} & \textbf{MinTL} \\ 
\hline
Flan-T5-Large & 1 & 1e-3 & 8 & 32 & 1e-3 & 512 & 400 & 8 \\ 
\hline
Biobart-V2-Large & 1 & 1e-3 & 8 & 32 & 1e-3 & 512 & 400 & 8 \\ 
\hline
\end{tabular}
\end{table*}

The performance of different Seq2Seq Models using LoRA and Fine-tuning can be found in Table \ref{tab:taskb_summary_comparison}
\begin{table*}[h]
    \centering
    \caption{SubTask B - Section Text Summarization Comparison}
    \label{tab:taskb_summary_comparison}
    \begin{tabular}{|l|l|l|}
    \hline
        \textbf{Base Architecture} & \textbf{LoRA-Score} & \textbf{Fine Tuning-Score} \\ \hline
        BioBart-V2-Large & 0.4310 & 0.2877 \\ \hline
        FLAN-T5-Large & 0.4276 & 0.1083 \\ \hline
    \end{tabular}
\end{table*}

\clearpage
\subsection{Standing of our team}
Our standings (in bold) for SubTask A - Section Header Classification is in Table \ref{tab:taska_section_header_classification_standings}. We omitted several teams from these standings and represent them by Ellipsis (\textbf{...}). This is done only to conserve space. 
\begin{table}[h]
    \centering
    \caption{SubTask A - Section Header Classification Standings}
    \label{tab:taska_section_header_classification_standings}
    \begin{tabular}{|l|l|l|l|}
    \hline
        \textbf{Team} & \textbf{Run} & \textbf{Accuracy} & \textbf{Rank} \\ \hline
        Cadence & run1 & 0.82 & 1 \\ \hline
        ... & ~ & ~ & ~ \\ \hline
        \textbf{SuryaKiran} & \textbf{run1} & \textbf{0.735} & \textbf{8} \\ \hline
        ... & ~ & ~ & ~ \\ \hline
        SSNSheerinKavitha & run1 & 0.14 & 23 \\ \hline
    \end{tabular}
\end{table}

Our standings (in bold) for SubTask B - Summarization is in Table \ref{tab:taskb_section_text_summarization_standings}
\begin{table}[h]
    \centering
    \caption{SubTask B - Section Text Summarization Standings}
    \label{tab:taskb_section_text_summarization_standings}
    \begin{tabular}{|l|l|l|l|l|l|l|}
    \hline
        \textbf{Team} & \textbf{Run} & \textbf{Rouge1} & \textbf{Bertscore\_F1} & \textbf{Bleurt} & \textbf{Aggregate\_score} & \textbf{Rank} \\ \hline
        \textbf{SuryaKiran} & \textbf{run3} & \textbf{0.4398} & \textbf{0.7231} & \textbf{0.5567} & \textbf{0.5732} & \textbf{1} \\ \hline
        ... & ~ & ~ & ~ & ~ & ~ & ~ \\ \hline
        \textbf{SuryaKiran} & \textbf{run2} & \textbf{0.4209} & \textbf{0.7137} & \textbf{0.5423} & \textbf{0.5590} & \textbf{5} \\ \hline
        \textbf{SuryaKiran} & \textbf{run1} & \textbf{0.4056} & \textbf{0.7109} & \textbf{0.5324} & \textbf{0.5496} & \textbf{6} \\ \hline
        ... & ~ & ~ & ~ & ~ & ~ & ~ \\ \hline
        SKKU-DSAIL & run1 & 0.2603 & 0.5929 & 0.5305 & 0.4612 & 13 \\ \hline
    \end{tabular}
\end{table}
\clearpage

\bibliography{sample-ceur}

\end{document}